\def\year{2022}\relax
\documentclass[letterpaper]{article} 
\usepackage{aaai22}  
\usepackage{times}  
\usepackage{helvet}  
\usepackage{courier}  
\usepackage[hyphens]{url}  
\usepackage{graphicx} 
\usepackage{natbib}  
\usepackage{caption} 
\DeclareCaptionStyle{ruled}{labelfont=normalfont,labelsep=colon,strut=off} 
\usepackage{algorithm}
\usepackage{algorithmic}

\usepackage{amsmath}
\usepackage{amssymb}
\usepackage{multirow}
\usepackage{booktabs}
\usepackage{xspace}

\DeclareRobustCommand{\eg}{e.g.\xspace}
\DeclareRobustCommand{\ie}{i.e.\xspace}
\DeclareRobustCommand{\etc}{etc.\xspace}

\DeclareRobustCommand{\vs}{vs.\xspace}
\DeclareRobustCommand{\pa}{$\mathcal{R}$Par\xspace}
\DeclareRobustCommand{\ox}{$\mathcal{R}$Oxf\xspace}
\DeclareRobustCommand{\dis}{$\mathcal{R}$1M\xspace}
\def\name{InsCLR\xspace}
\renewcommand{\thefootnote}{\fnsymbol{footnote}}

\usepackage{newfloat}
\usepackage{listings}
\floatstyle{ruled}
\newfloat{listing}{tb}{lst}{}
\floatname{listing}{Listing}
\title{InsCLR: Improving Instance Retrieval with Self-Supervision}

\author {
    Zelu Deng\textsuperscript{\rm 1}$^{*}$,
    Yujie Zhong\textsuperscript{\rm 2}$^{*}$,
    Sheng Guo\textsuperscript{\rm 3},
    Weilin Huang\textsuperscript{\rm 4}$^\dagger$
}
\affiliations {
    \textsuperscript{\rm 1}Dmall Inc.,
    \textsuperscript{\rm 2}Meituan Inc.,
    \textsuperscript{\rm 3}MYBank, Ant Group,
    \textsuperscript{\rm 4}Alibaba Group \\
    zelu.deng@dmall.com, zhongyujie@meituan.com, \{guosheng.guosheng, weilin.hwl\}@alibaba-inc.com
}

\usepackage{bibentry}

\begin{document}

\maketitle

\begin{abstract}
\vspace{-1mm}
This work aims at improving instance retrieval with self-supervision.
We find that fine-tuning using the recently developed self-supervised (SSL) learning methods, such as SimCLR and MoCo, fails to improve the performance of instance retrieval.
In this work, we identify that the learnt representations for instance retrieval should be invariant to large variations in viewpoint and background etc., whereas self-augmented positives applied by the current SSL methods can not provide strong enough signals for learning robust instance-level representations. 
To overcome this problem, we propose InsCLR, a new SSL method that builds on the \textit{instance-level} contrast, to learn the intra-class invariance by dynamically mining meaningful pseudo positive samples from both mini-batches and a memory bank during training. 
Extensive experiments demonstrate that InsCLR achieves similar or even better performance than the state-of-the-art SSL methods on instance retrieval. Code is available at https://github.com/zeludeng/insclr.
\vspace{-2mm}

\end{abstract}

\newcommand\blfootnote[1]{%
\begingroup 
\renewcommand\thefootnote{}\footnote{#1}%
\addtocounter{footnote}{-1}%
\endgroup 
}
{
	\blfootnote{
	 $^*$Equal contributions. 
	 $^\dagger$Corresponding author.
	 }
}

\section{Introduction}\label{sec:intro}

Large-scale instance-level image retrieval (also known as particular object retrieval) has been studied for over two decades. 
Given an image with a query object, the goal is to retrieve all the images containing the query object in a large corpus of images.
A key challenge is to design or learn image-level descriptors for the accurate search.
Recently, several works show that fine-tuning the pretrained network on large-scale instance-retrieval datasets can significantly improve the performance~\cite{gordo2016deep,noh2017large,weyand2020google}. 
However, annotating large-scale data is time-consuming and requires huge human labour. 
Alternatively, image labels can be generated by using the reconstructed 3D models obtained from a traditional retrieval system \cite{radenovic2016cnn,radenovic2018fine}, and the advanced Structure-from-Motion (SfM) pipeline~\cite{schonberger2016structure}.
It can be considered as training networks without human annotation, but using supervision information generated from other  computer vision systems which are also expensive to implement.

\begin{figure}[t!]
\centering
\includegraphics[width=0.99\columnwidth]{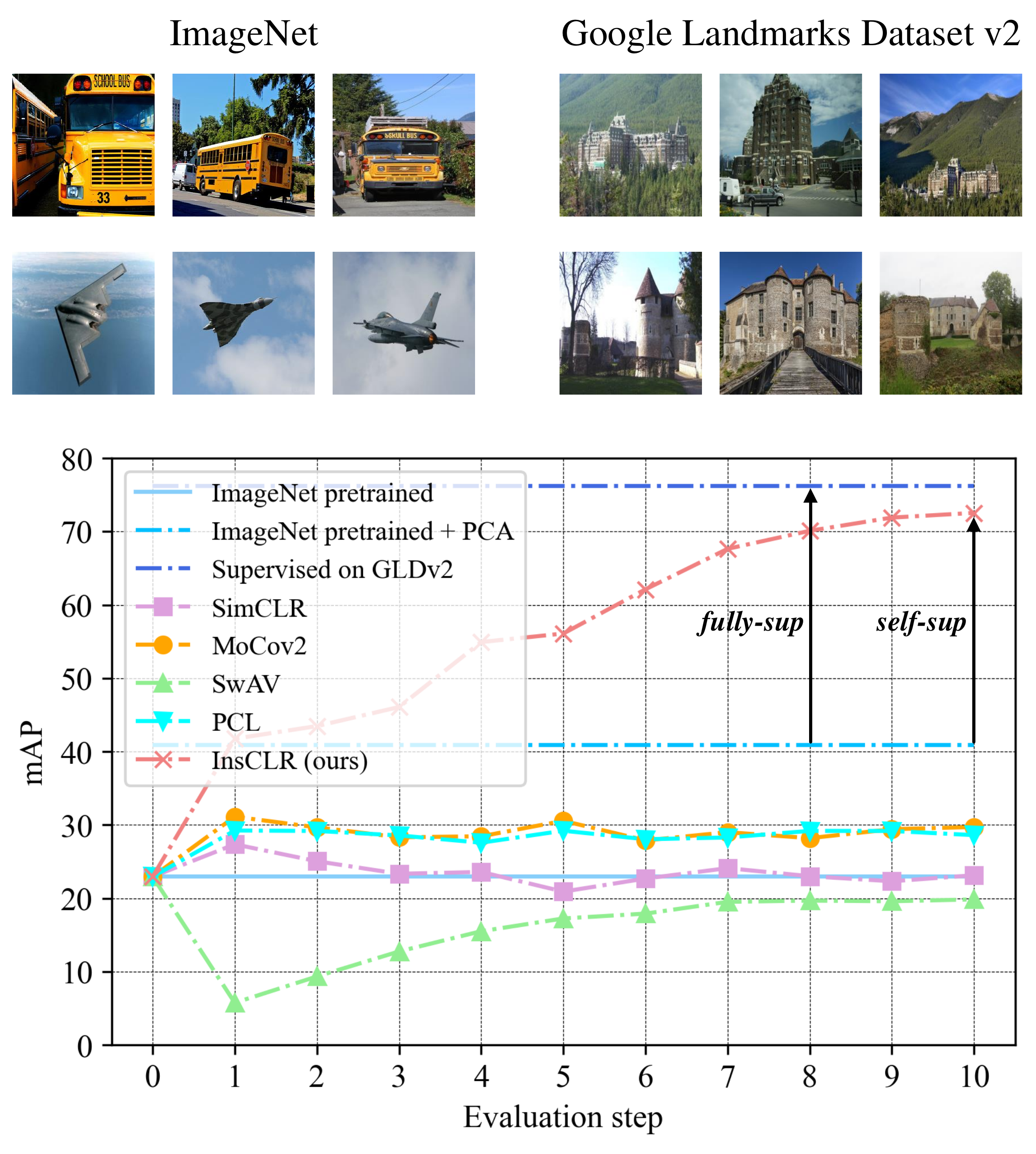}
\vspace{-4mm}
\caption{ {\textbf{Top}: example images from ImageNet  and Google Landmarks Dataset v2  (GLDv2).
In GLDv2, images of the same class (\ie those in the same row) have much larger variations in viewpoint and background.
\textbf{Bottom}:  the performance of instance retrieval on \ox (medium protocol), with a number of self-supervised learning methods, including
SimCLR, MoCov2, PCL, SwAV and our \name.  ResNet-101 pretrained on ImageNet is used for all methods, and then self-supervised learning is performed on GLDv2. 
} 
} 
\vspace{-6mm}
\label{fig:teaser}
\end{figure}

In this work, we aim to explore a more generic setting: can we improve the retrieval quality by using images only, without human annotations  or other computer vision systems?
Formally, the goal is to close the performance gap between a network trained on a general-purpose dataset (\ie \emph{ImageNet}) and a network fine-tuned with \emph{full-supervision} on instance-retrieval datasets, by using unlabeled images only. 
Notably, this work focus on the fine-tuning stage instead of training from scratch.
A recent approach described in ~\cite{iscen2018mining} works with such assumptions. It constructs positive and negative samples using a manifold similarity~\cite{iscen2017efficient} computed from a similarity graph which is built on Euclidean distance. However, it focuses on designing an offline preprocessing of the images to generate the corresponding image labels, and the training procedure remains the same as the standard supervised training.
In this work, we go beyond such offline preprocessing, and provide a more dynamic solution that generates self-supervised learning signals in an online manner during training, with minimal offline computation.

Recent self-supervised learning (SSL) methods, such as MoCo~\cite{he2020momentum} and SimCLR~\cite{chen2020simple}, train networks by learning instance discrimination between self-image (or self-augmented image) and other images.
However, for the task considered in this work, \textit{we find that simply applying these state-of-the-art SSL methods to learn image-level representations for instance retrieval is far from ideal.} As Figure~\ref{fig:teaser} shows,  performing self-supervised learning with SimCLR or MoCov2~\cite{chen2020improved} with an ImageNet-pretrained network on
the recently released Google Landmarks Dataset v2~\cite{weyand2020google} (without using ground-truth labels) can not obtain the expected performance on the public benchmark  for instance retrieval: revisited Oxford~\cite{radenovic2018revisiting}. 

As Figure~\ref{fig:teaser} (top) shows, different from ImageNet, objects in instance retrieval may have large variance in viewpoint, background clutter, occlusion and illumination conditions \etc.  
Therefore, an important capability for instance retrieval is to learn strong object representations that are robust to the large intra-class variation, and to focus on discriminating object instances rather than images.
However, existing image-level SSL methods (such as MoCov2 and SimCLR) can not fully explore the intra-class information which is particularly useful to instance retrieval in datasets like GLDv2.

Consequently, we introduce an \textit{instance-level} SSL method that learns to capture the abovementioned properties from instance-retrieval datasets.
The proposed method explores contrastive learning signals explicitly from intra-class pairs by mining cross-image  pseudo positives from both the mini-batches and a memory bank along the training. It encourages the model to pull the images of the same class but having different viewpoints or backgrounds closer in the feature space. The mined positives provide much more meaningful learning signals than the self-augmented image pairs, particularly on learning robust intra-class representations.
The proposed method is code-named \name.
We make the following contributions:

-- We identify the limitation of state-of-the-art image-level SSL methods such as  MoCov2 and SimCLR, and propose \name for \textit{instance-level SSL} 
which is able to learn strong instance representations robust to large intra-class variance.

-- To build meaningful instance-level contrastive information, we propose new algorithms to dynamically mine pseudo positives from both mini-batches and the memory bank in the contrastive learning framework.

-- Extensive experiments across three public benchmarks (revisited Oxford, Paris and INSTRE) demonstrate that the proposed \name surpasses all other self-supervised methods, and even outperforms many recent supervised methods.
In particular, as Figure~\ref{fig:teaser} shows, our \name achieves 73.1 mAP on the revisited Oxford (medium), \textit{significantly} closing the gap between the unsupervised fine-tuning and the best-performing supervised counterpart~\cite{weyand2020google} (76.2 mAP) on instance retrieval.

\vspace{-1mm}

\section{Related Work}

\begin{figure*}[t]
\centering
\includegraphics[width=0.99\linewidth]{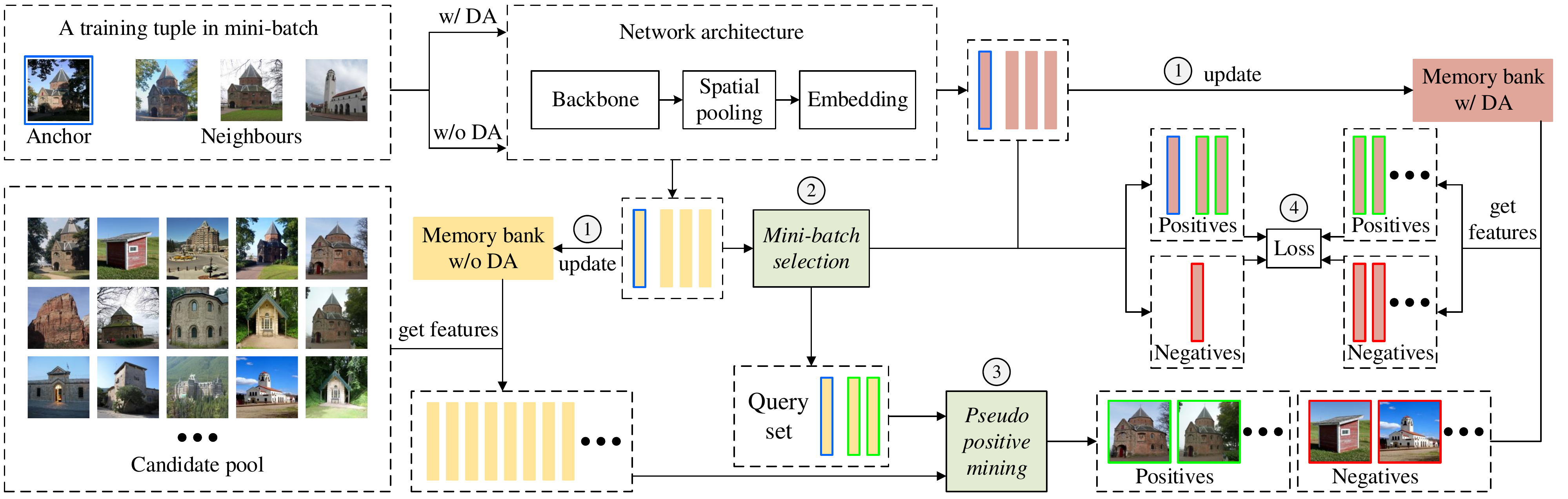}
\vspace{-1mm}
\caption{ 
\textbf{Overview of \name.}
During training, each image is fed to the network in two forms: with and without random data augmentations (DA). The former (red) contributes to the back-propagation, while the latter (yellow) is used for the robust positive mining. 
\textbf{Step 1}: the images are encoded by the network, and the output features are first used to update those in the corresponding memory banks.
\textbf{Step 2}: positives are selected within the mini-batch (Section~\ref{subsec:batch}). The selected positive features without DA form a query set.
\textbf{Step 3}: with the query set, pseudo positives are mined from the memory features corresponding to the candidate pool of the anchor (Section~\ref{subsec:memory}). The features that are not selected as positives become negatives.
\textbf{Step 4}: a contrastive loss is computed based on the anchor, the mined positives and negatives (with DA).
}
\vspace{-6mm}
\label{fig:method}
\end{figure*}

\paragraph{Image representations for instance-level image retrieval.}
In image retrieval, representing images with image-level (\ie global) features is particularly favoured in practice due to its run-time efficiency. In the era of deep learning, global features can be generated by aggregating CNNs features~\cite{babenko2015aggregating,tolias2015particular,arandjelovic2016netvlad,gordo2016deep,radenovic2016cnn,tolias2016image,noh2017large,radenovic2018fine}.
Apart from global features, local features are also used to perform spatial verification~\cite{philbin2007object,noh2017large,cao2020unifying}, which incorporates the geometric information of objects and results in a more reliable matching. 
In this work, we focus on learning global image descriptors with self-supervision due to its simplicity, and leave local descriptors for future work.

\vspace{-2mm}
\paragraph{Self-supervised representation learning.}
Recently, prominent performance in image classification is achieved by contrastive learning~\cite{oord2018representation,he2020momentum,chen2020simple,chen2020improved, caron2020unsupervised}. In particular, MoCo~\cite{he2020momentum,chen2020improved},  SimCLR~\cite{chen2020simple} and SwAV~\cite{caron2020unsupervised} further reduce the performance gap between self-supervised networks and fully-supervised networks.
Different from MoCo and SimCLR that learn by the image discrimination,  SCAN~\cite{van2020scan} and InterCLR~\cite{xie2020delving}
reveal that more important semantic information can be explored across images. 
Recent works, such as PCL~\cite{li2020prototypical} and SwAV~\cite{caron2020unsupervised}, were developed to learn intra-class information implicitly by assigning similar images to same prototypes/clusters. 
In this work, we show that the intra-class information can be better explored by explicitly finding cross-image positive pairs with a pairwise learning paradigm.

\vspace{-2mm}
\paragraph{Feature memory bank.}
In contrastive learning framework, memory banks can be used in both supervised~\cite{li2019memory,wang2020cross} and unsupervised learning~\cite{he2020momentum,chen2020improved} with different motivations. 
Different from them, we propose to mine both positive and negative samples in the memory for unsupervised learning, which has never been explored before.

\vspace{-1mm}

\section{The proposed \name}
\vspace{-1mm}

In this section, we present details of the proposed \name which is able to learn strong image representations for instance retrieval by mining pseudo positives and negatives in a self-supervised manner.  
We start from an overview of the method (Section~\ref{subsec:overview}), including a formal definition of the task, the network architecture and the setup for training samples. Then, we describe details of mining positives in mini-batches (Section~\ref{subsec:batch}) and the memory bank (Section~\ref{subsec:memory}). The training loss is defined in Section~\ref{subsec:loss}.

\subsection{Overview}
\label{subsec:overview}
\paragraph{Problem definition.} We follow the line of work that represents images by global features~\cite{babenko2015aggregating,tolias2015particular,gordo2016deep,cao2020unifying} extracted from CNNs. Image retrieval is then performed by computing a cosine similarity between a query image  and a  set of gallery images in the feature space: $S_{ij}$ = $\cos(f_i, f_j)$, where $f$ denotes the image feature extracted by the CNN.
In this case, the retrieval quality entirely depends on image-level representations computed from CNNs.
Given an ImageNet-pretrained network and an instance-retrieval-oriented dataset, the objective of this work is to learn strong image representations for instance retrieval by using the dataset in a self-supervised learning (SSL) manner.
As discussed in Section~\ref{sec:intro}, the state-of-the-art SSL methods are mainly designed for image classification, and fail to capture the large intra-class invariance (such as viewpoint, background \etc) for instance retrieval.
In this work, we propose \name to close this gap by mining informative cross-image positives during training, and manage to match the performance of fully-supervised methods.

\vspace{-1mm}
\paragraph{The learning framework of \name.}
As illustrated in Figure~\ref{fig:method}, \name mainly consists of a network to encode images, memory banks to store image features, and the proposed methods to mine pseudo positives from both mini-batches and the memory bank.
For each anchor image, the mined positives as well as the negatives are used to compute a contrastive loss for training the network.
We briefly describe the adopted network architecture and  our training sample configuration in the following.

\vspace{-1mm}
\paragraph{Network architecture.}
To make a fair comparison, we adopt s simple network architecture to produce image-level features. As shown in Figure~\ref{fig:method} (top-middle), it consists of three components: a backbone network, a spatial pooling layer and an embedding module. Details on the architecture are presented in Section~\ref{subsec:imp}.

\vspace{-1mm}
\paragraph{Memory bank.}
We leverage memory banks to store a large amount of sample representations during training, which provide more diverse yet meaningful hard samples apart from those in mini-batches.
Different from previous image-level SSL methods~\cite{wu2018unsupervised,he2020momentum} that regard all the features in the memory bank as  negative samples, we propose to mine pseudo positive samples from the memory bank. 
Although an additional momentum encoder~\cite{he2020momentum} can be used to alleviate the problem of inconsistency between the features in the memory, we only maintain one encoder (similar to \cite{wang2020cross}) due to its efficiency and simplicity.

\vspace{-1mm}
\paragraph{Setup for training samples.}
SCAN~\cite{van2020scan} simply collects positives for each mini-batch by computing the nearest neighbours of an anchor image, using a pretrained model. The nearest neighbours are computed offline, and are fixed during the whole training process. 
This inspired us to first compute the nearest neighbours for each image using a pre-trained model, which can be used in the subsequent step as prior knowledge  for dynamically selecting positives during training. 
Specifically, we construct a pre-computed candidate pool for each image, which contains the potential positives as well as the hard negatives (with a high similarity).
The candidate pool for each image is obtained by computing its $P$ nearest neighbours in the whole training set using only the global features extracted by ImageNet-pretrained networks in an offline manner. 
A training tuple is formed by an anchor image with its $N_b$ top-ranked images from its candidate pool. 
Anchors are randomly selected from the whole training set.
A mini-batch is then constructed by multiple such tuples, \eg 16 training tuples with $N_b = 3$ form a mini-batch of 64.

\vspace{-2mm}
\subsection{Positive Selection in Mini-batches}
\label{subsec:batch}
\vspace{-1mm}

To identify more informative positives during training, we investigate various approaches to dynamically collect positives from each tuple, which are described as follows.

We consider taking all the $N_b$ images in the tuple as positives, referred as \emph{nn} (similar to~\cite{van2020scan}), as a baseline. We then investigate four threshold-based strategies to select the positives from the $N_b$ images. Given a threshold $T_b$, the four methods are defined as follows.

(1) \textbf{Augmented similarity.} We adopt a threshold to select the positives, \ie computing feature similarities between the anchor and $N_b$ neighboring images in the training tuple, and only considering the images with a similarity over the predefined threshold as pseudo positives: $S_{ij}^{DA}$ $>$ $T_b$.
However, this similarity is highly unreliable since it is computed using the images after random augmentations.
(2) \textbf{Unaugmented similarity.} To overcome this limitation, the second strategy is to feed the original images without any augmentation to the networks, and apply the threshold on their similarities: $S_{ij}^{w/oDA}$ $>$ $T_b$. Note that the unaugmented version is only used for similarity computation, and does not contribute to training loss.
(3) \textbf{Sample-relative similarity.} A universal threshold may not work reliably to all anchor images.
Some classes may have smaller intra-class variance, and thus require larger thresholds. We further develop the third strategy that selects the positives by using sample-relative similarities. Namely, the similarities are scaled by dividing by the largest similarity in each training tuple: $S_{ij}^{rel}$ $>$ $T_b$.
(4) \textbf{Multi-scale similarity.} Lastly, based on the second strategy, we intend to improve the similarity by feeding the unaugmented images with multiple scales, which is the fourth strategy: $S_{ij}^{ms}$ $>$ $T_b$.

With these mining approaches, we can select pseudo positives dynamically within each mini-batch.
For example, a training tuple with $N_b = 7$ may have 3 images selected as positives, based on one of the proposed methods, and the other 4 are then regarded as negatives.
We empirically compare the four strategies in Section~\ref{subsec:ablation}.

\vspace{-2mm}
\subsection{Mining Positives from Memory Bank}
\label{subsec:memory}
\vspace{-0mm}

\paragraph{Benefits.}
Apart from learning the discrimination between positives and negatives in mini-batches, we also wish to collect more positives from the memory bank. The benefit of finding positives from the memory bank is two-fold. 
First, mining more positives from the memory bank will encourage the model to pull potential positives closer, instead of pushing them away by default (\ie considering them as negatives in the memory).
This sets our method apart from image-level SSL like MoCo~\cite{he2020momentum}  or SimCLR~\cite{chen2020simple}.
Second, by excluding the selected positives, the rest of the images in the candidate pool are considered as hard negatives, since they often have high offline similarities with the anchor image (comparing to other images in the dataset).

\begin{figure}[t]
\centering
\includegraphics[width=0.99\columnwidth]{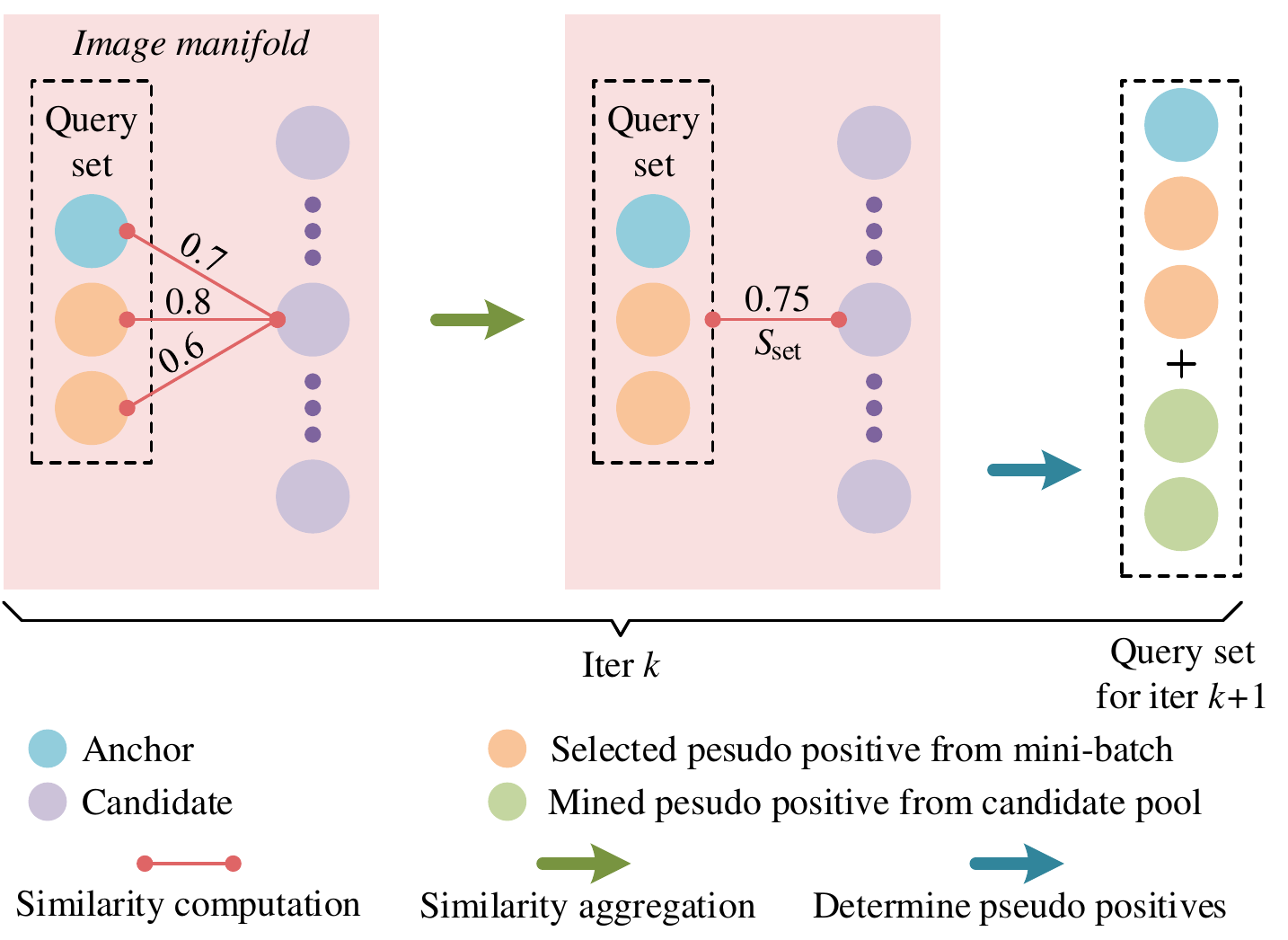}
\caption{
\textbf{Mining positives in the memory bank.}
The similarities $S_{set}$ between each image in the candidate pool and the whole query set are computed by pairwise similarity computation followed by aggregation. 
The positive selection is based on $S_{set}$ and the mined images become part of the new query set.
} 
\vspace{-4mm}
\label{fig:memory}
\end{figure}

\vspace{-2mm}
\paragraph{Mining with query sets.}
The selected positives within the mini-batch are assumed to be of the same class as the anchor, with a high confidence. Therefore, we can consider the anchor image and its selected positives from the mini-batch as a query set, and then cast the task of mining positives from the memory bank into a retrieval problem with \emph{a set of} query images, instead of a single query image.
To this end, we propose a new algorithm that can effectively explore the underlying image relation on-the-fly during training. 
This procedure is presented in the bottom part of Figure~\ref{fig:method}.
Crucially, the whole mining process should be performed on the image features extracted without any random augmentation. 
Otherwise, we found the training can degrade significantly. 
Hence, two memory banks are used. The one without augmentations is for mining, while the other one is for representation learning.
To be specific, our method consists of two steps: similarity computation and aggregation, which are performed on every training tuple in a mini-batch.
The method is shown in Figure~\ref{fig:memory}.

\vspace{-1mm}
\paragraph{Step 1: similarity computation.}
To make use of every query at hand, we first compute the similarity between the features of each query image  and that of memory images.
Each image in the pool now has $N_p$ similarity scores, where $N_p$ denotes the size of the query set $Q$.
As an optional step, we can disregard the similarity scores below a threshold. 
Intuitively, it is possible to have an image which may not look similar to all the images of the same class, \eg even though images contain the same object, they may not have a low global similarity due to different viewpoint and background clutter \etc 
Mathematically, we have:
\begin{flalign}
\vspace{-1mm}
\begin{split}
S(i, Q) = \phi(S(i, q_1), S(i, q_2), \dots, S(i, q_{N_p})),
\end{split}
\vspace{-1mm}
\end{flalign}
where $\phi$ is the optional discarding step and $q \in Q$.

\vspace{-1mm}
\paragraph{Step 2: similarity aggregation.}
For each image in the candidate pool, we measure its similarity to the whole query set by $S_{set}$.  $S_{set}$ is obtained by aggregating its similarities to each image in the query set.
In this work, two aggregation functions are considered: average and maximum:
\begin{flalign}
\vspace{-1mm}
\begin{split}
S_{set} = \psi(S(i, Q)),
\end{split}
\vspace{-1mm}
\end{flalign}
where $\psi$ is the aggregation function.
We then re-rank the images in the candidate pool based on $S_{set}$. Given the re-ranked of candidate images, we can determine the pseudo positives either using a threshold $T_m$ or the top-$k$ rule. 
The mined positives can be added to the query set $Q$ and we can repeat the above two steps for several times to gather more positives if desired.
Apart from the mined positives, the rest of the candidate pool is then used as negatives.

\begin{figure*}[t]
\centering
\includegraphics[width=0.90\linewidth]{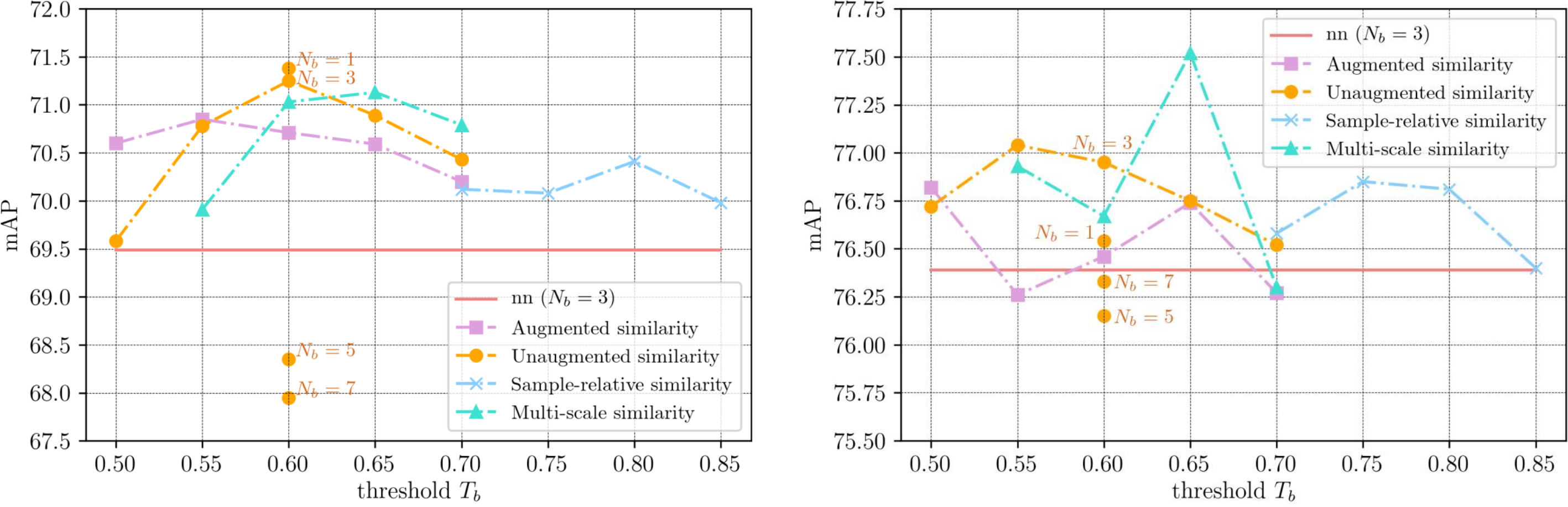}
\vspace{-3mm}
\caption{ 
\textbf{Ablation on the positive selection in mini-batches.}
The results are obtained on \ox (left) and \pa (right) with medium protocol.
\emph{nn} considers all the $N_b$ images in each training tuple to be positives, without selection.
Notably, the results of augmented-image-based similarity are not shown in the plots: 35.7 mAP and 39.5 mAP for \ox and \pa, respectively.
}
\vspace{-5mm}
\label{fig:mini-batch}
\end{figure*}

\vspace{-1mm}
\paragraph{Efficient online graph traversal.}
The proposed method can be considered as an online graph traversal in the image manifold of the query set and the candidate pool. It has two advantages comparing to the standard offline graph traversal methods like~\cite{iscen2017efficient,iscen2018mining,chang2019explore}.
\textbf{First}, it can be applied during training without introducing much computational overhead. It avoids computing the similarities among images in the candidate pool. Instead, similarities are only measured between the candidate pool and the query set. 
\textbf{Second}, it is an online approach and the image manifolds can become more reliable along the training, in contrast to~\cite{iscen2018mining} in which the image labels are generated in the fixed image manifolds before training.

\vspace{-2mm}
\subsection{Loss}
\label{subsec:loss}
We adopt a simple contrastive loss~\cite{hadsell2006dimensionality}. 
Concretely, the loss function $L$ for a training tuple is:
\begin{flalign}
\vspace{-2mm}
\begin{split}
    L = \frac{1}{N_{p}}\sum_{i=1}^{N_{p}}\left [  \sum_{y_i \neq y_j}^{N_{iter}}S_{ij} - \sum_{y_i = y_j}^{N_{iter}}S_{ij}\right ],
\label{eq:loss}
\end{split}
\vspace{-2mm}
\end{flalign}
\noindent where $N_p$ is the size of query set after memory mining.
For a training iteration, $N_{iter}$ is a collection of all images in the mini-batch and the candidate pool for the anchor image.
$y_i \neq y_j$ indicates a negative pair while $y_i = y_j$ denotes a positive pair. 
$S_{ij}$ is the cosine similarity between features.

\vspace{-2mm}

\section{Experiments and Results}
\label{sec:exp}

\subsection{Implementation Details}
\label{subsec:imp}

\paragraph{Network architecture.}
Following~\cite{weyand2020google}, ResNet-101~\cite{he2016deep} is used as the backbone network, and only the first four convolutional blocks are kept. The channel size of the backbone output feature map is 2048. 
For the spatial pooling layer, we adopt the Generalized Mean pooling~\cite{radenovic2018fine} (GeM) with the parameter $p$ fixed to be 3. 
Similar to~\cite{gordo2017end,cao2020unifying}, we use a fully-connected layer as the embedding module with an output dimension of 2048, without a careful tuning.

\vspace{-1mm}
\paragraph{Training details.}
The training data is a subset of GLDv2~\cite{ozaki2019large}.
The dataset contains 1.2M images from 27k landmarks. 
Unless specified, the size of the offline-computed candidate pool $P$ is set to be 500 for every image, and $N_{b}$ is set to be 3 for all networks. 
More details regarding the training and the evaluation protocol (including feature extraction) refer to the supplementary material.

\subsection{Ablation study}
\label{subsec:ablation}
We conduct ablation study on two public benchmarks: Oxford and Paris with revisited annotations~\cite{radenovic2018revisiting}, denoted by \ox and \pa, respectively.

\paragraph{Candidate pool size.}
We empirically find that the performance increases along with $P$ until around 250, after which the performance improves marginally. This is probably because only the hard negatives with similarity higher than 0.4 contribute to the loss, and increasing the size of the pool over 200 brings minimal hard negatives.

\vspace{-1mm}
\paragraph{Positive selection in mini-batches.}
As Figure~\ref{fig:mini-batch} shows, the proposed pseudo positive selection strategies improve the performance over the nearest-neighbour baseline \emph{nn} in general. In particular, applying an absolute threshold on the unaugmented similarity performs better than others, only beaten by its multi-scale variant on \pa. 
Furthermore, the similarity based on the augmented images is highly unreliable and harms the retrieval performance severely. 
This proves the importance of feeding a version without data augmentation for each image during training. 
Based on the unaugmented similarity, we experiment on several values of $N_b$ (\ie 1, 3, 5, 7), and find that $N_b=3$ is the optimal choice.
In the rest of the experiments, a threshold of $T_b$ = 0.65 with the unaugmented similarity is adopted, with $N_b$ = 3.

\begin{table}[t]
\centering
{ \small
\begin{tabular}{l c c c c c c }
 \multirow{2}{*}{Method}  & \multirow{2}{*}{Batch} & \multirow{2}{*}{Memory} & \multicolumn{2}{c}{ \small Medium} & \multicolumn{2}{c}{ \small Hard} \\ \cmidrule[0.5pt]{4-7} 
 
 & && \multicolumn{1}{c}{\large \vphantom{M} \scriptsize \ox } & \multicolumn{1}{c}{\scriptsize \pa}  & \multicolumn{1}{c}{\scriptsize \ox}  & \multicolumn{1}{c}{\scriptsize \pa}  \\ \hline

\multicolumn{3}{l}{ImageNet}  &  23.0  & 52.0 &  6.5  & 25.9 \\ 
\multicolumn{3}{l}{ImageNet + PCA}  & 40.9   & 65.6 &  30.5  & 41.7 \\ 
\multicolumn{3}{l}{ImageNet + PCA$^{*}$}  & 45.0   & 70.7 & 17.7   &  48.7 \\ 
\multicolumn{3}{l}{Supervised-Arc$^{\dagger}$} & 76.2 & 86.8 & 55.1 & 72.5 \\
\cmidrule[0.5pt]{1-3} \cmidrule[0.5pt]{4-7}

A & \emph{nn} & - &   57.6 & 68.1 &  30.3 & 43.7  \\
B & \emph{nn} & neg. &  61.0  & 73.4 &  35.4  &  52.6 \\ 
C & ours & neg. &  65.2  & 75.1 & 39.9   & 55.7 \\ 
D-anchor  & ours & anc. &  67.4  & 75.6 &   42.2  & 56.5 \\ 
D & ours & ours &   73.1  & 77.6  &  48.3  & 59.5 \\   

\end{tabular}
}
\vspace{-3mm}
\caption{\textbf{Ablation on pseudo positive mining.} All methods use ResNet-101 with GeM pooling.
\emph{nn} denotes that taking all the images in the training tuple as positives without selection. \emph{neg.} means that all of the $10^{5}$ features randomly sampled from the memory bank  are considered as negatives.
{*} is from~\cite{radenovic2018revisiting}, and $\dagger$ is from ~\cite{weyand2020google} trained with ArcFace loss~\cite{deng2019arcface}.
}
\label{tab:1234}
\vspace{-5mm}
\end{table}

\vspace{-1mm}
\paragraph{Performance gain of each component.}
We compare our method with several baselines and show the performance gain brought by each proposed component in Table~\ref{tab:1234}.
Firstly, the ImageNet-pretrained network can be seen as a lower bound. As reported in~\cite{radenovic2018revisiting}, applying a PCA/whitening on the features can significantly increase the performance. 
Notably, the rest of the results in Table~\ref{tab:1234} do not involve PCA post-processing. 
Training with offline-computed $nn$ positives in the mini-batch (denoted by method \emph{A}) raises the performance to a relatively high level, \ie above 60 with medium setup. 
Method \emph{B} is method \emph{A} with a memory bank in which features are only regarded as negatives. This gives notable gains in both \ox and \pa. 
Method \emph{C} replaces the $nn$ with the proposed pseudo positive selection, which improves the mAP by around 4 and 3 points in \ox and \pa, respectively.
If the proposed memory mining is adopted (denoted by method \emph{D}), the performance is further boosted prominently. 
However, if we perform the mining in the memory only using the anchor image itself (\ie without the selected positives in the training tuple), the performance drops significantly. This validates the importance of mining with a query set.

\vspace{-2mm}
\paragraph{Details in memory mining.}

In Table~\ref{tab:memory}, we compare the different options in the design of memory mining described in Section~\ref{subsec:memory}. 
For the similarity aggregation methods, \emph{mean} and \emph{max}
perform better in \ox and \pa, respectively. 
As the selection strategy, \emph{topk} performs better than \emph{threshold}.
This is in contrast to the case in mini-batches, where \emph{threshold} is better. It is because the potential positives from mini-batches are the \emph{top-ranked} images from the candidate pool, which are very likely to be easy positives. Hence, the taking thresholds on the similarity is relatively reliable. Whereas in the memory bank, mining is more difficult and similarities become less reliable.
In terms of sparsity, the performance seems not sensitive. This is probably because the size of the query set is relatively small. 
The last conclusion to draw from Table~\ref{tab:memory} is that the mAP increases with the mining iterations, while bringing negligible computational overhead.
In the rest of the experiments, \emph{avg} and \emph{topk} are adopted with 4 iterations in the mining.

\begin{table}[t]
\centering
{\small
\begin{tabular}{l l c c c c }
 \multirow{3}{*}{Aggre.}  & \multirow{3}{*}{Selection} & \multicolumn{2}{c}{ \small Medium} & \multicolumn{2}{c}{ \small Hard} \\ \cmidrule[0.5pt]{3-4} \cmidrule[0.5pt]{5-6}
 
& & \multicolumn{1}{c}{\large \vphantom{M} \small \ox } & \multicolumn{1}{c}{\small \pa}  & \multicolumn{1}{c}{\small \ox}  & \multicolumn{1}{c}{\small \pa}  \\ \hline

avg & topk &  71.3  & 77.0  &  45.5  & 58.3 \\ 
avg & topk w/ spa. &  70.4  & 77.1 &  44.8  & 58.8  \\
avg & threshold  & 68.4  & 76.3 &  42.0  &  57.4 \\
max & topk &  70.6  & 77.2  &   44.5 & 58.7 \\ 
max & topk w/ spa. & 70.4    & 77.2  & 44.0   & 58.5  \\ 
max & threshold &  61.1  & 71.5 &  34.9  & 47.6 \\ \hline

avg & topk 1$\times$ &  69.5  & 76.2  &  42.8  & 57.2 \\ 
avg & topk 4$\times$ &  73.1  & 77.6  &  48.3  & 59.5 \\

\end{tabular}
}
\vspace{-1mm}
\caption{\textbf{Ablation on mining positives in the memory bank.} 
\emph{topk} denotes taking the top $k=5$ images in the ranked candidate pool.
$T_m$ = 0.6 for \emph{threshold}.
\emph{spa.} refers to retaining only the similarity scores above  0.6 to enforce the sparsity. 
\emph{1$\times$} or \emph{4$\times$} means that the mining is performed for one or four iterations (otherwise two iterations).
}
\label{tab:memory}
\vspace{-6mm}
\end{table}
\begin{table*}[t]
\centering
{\small
\begin{tabular}{l c c c c c c c c }
 \multirow{3}{*}{Method}  & \multicolumn{4}{c}{ \small Medium} & \multicolumn{4}{c}{ \small Hard} \\ \cmidrule[0.5pt]{2-5} \cmidrule[0.5pt]{6-9}
 
 & \multicolumn{1}{c}{\large \vphantom{M} \small \ox } & \multicolumn{1}{c}{\small +\dis} & \multicolumn{1}{c}{\small \pa} & \multicolumn{1}{c}{\small +\dis} & \multicolumn{1}{c}{\small \ox} & \multicolumn{1}{c}{\small +\dis} & \multicolumn{1}{c}{\small \pa} & \multicolumn{1}{c}{\small +\dis} \\ \hline

\multicolumn{9}{c}{Supervised training} \\

R101 - GeM (ImageNet) \cite{radenovic2018revisiting}  & 45.0  & 25.6 & 70.7 & 46.2  & 17.7  & 4.7  & 48.7 & 20.3 \\ 
R101 - R-MAC \cite{gordo2017end} &   60.9  & 39.3 & 78.9 & 54.8 & 32.4 & 12.5 & 59.4 & 28.0 \\ 
R101 - GeM - AP \cite{revaud2019learning} &  67.5 &  47.5 &  80.1 &  52.5 &  42.8 &  23.2 &  60.5 &  25.1 \\
R101 - GeM - AP (GLD) \cite{revaud2019learning} &  66.3 & -- &  80.2 & -- &  42.5 & -- &  60.8 & -- \\
R101 - DELG \cite{cao2020unifying} & 73.2 & $54.8$ & 82.4 & 61.8 & 51.2 & 30.3 & 64.7 & 35.5 \\ 
R101 - GeM (GLDv2-clean) \cite{weyand2020google} & 76.2 & - &  86.8 & - & 55.1 &  - & 72.5 & - \\

DELF - ASMK + SP
\cite{radenovic2018revisiting} & 67.8  & 53.8 & 76.9 & 57.3 & 43.1 & 31.2 & 55.4 & 26.4 \\ 
DELF - R-ASMK + SP (GLD) \cite{teichmann2019detect}  & 76.0 &  64.0 &  80.2 &  59.7 &  52.4 &  38.1 &  58.6 &  29.4 \\

\hline
\multicolumn{9}{c}{Self-supervised training based on automatic annotation} \\ 

VGG16 - MAC~\cite{radenovic2016cnn} & 58.4 & 39.1 & 66.8 & 42.4 & 30.5 & 17.9 & 42.0 & 17.7 \\
R101 - GeM (GLD) \cite{radenovic2018fine} &  64.7  &  45.2 &  77.2  &  52.3 &  38.5 &  19.9 &  56.3 &  24.7 \\
R152 - GeM (GLD) \cite{radenovic2018fine} &  68.7  & -- &  79.7  & -- &  44.2 & -- &  60.3 & -- \\
R101 - GeM \cite{simeoni2019local} &  65.3 &  46.1 &  77.3 &  52.6 &   39.6 &  22.2 &  56.6 &  24.8 \\
R101 - GeM + DSM \cite{simeoni2019local} &  65.3 &  47.6 &  77.4 &  52.8 &   39.2 &  23.2 &  56.2 &  25.0 \\

\hline
\multicolumn{9}{c}{Self-supervised training} \\ 
DeepCluster~\cite{caron2018deep} &  29.8  & 9.8  & 49.1   & 13.3 & 9.0  & 0.9 & 26.0   & 3.2 \\
SimCLR~\cite{chen2020simple} & 22.2  & 9.4  &  50.5   & 14.4  &  6.3 & 2.2  &  19.6   & 1.1  \\
MoCov2~\cite{chen2020improved} & 27.3   & 11.0  &  65.1  & 17.4 & 6.1  & 0.8 & 38.4   & 3.2 \\
BYOL~\cite{grill2020bootstrap} & 11.0  & 1.9  & 28.4    & 3.7  &  2.3 & 0.1  &   8.8  & 0.2  \\
PCL~\cite{li2020prototypical} & 29.2   & 10.3  &  59.3    & 17.6  & 7.9   & 0.5  & 28.9    & 2.6  \\
SwAV~\cite{caron2020unsupervised} &  19.9  & 7.1  & 38.5     &  10.4 &  3.7  &  0.1 &  10.5   & 0.4  \\

\name (ours) &   73.1 &  56.2 &  77.6  & 56.7 &  48.3 & 29.6 &  59.5  & 29.2 \\

\end{tabular}
}
\vspace{-2mm}
\caption{\textbf{Comparison to state-of-the-art methods on large-scale retrieval.} 
\emph{Automatic annotation} means additional computer vision systems are used to annotate the images before training.
\emph{SP} refers to the spatial verification using local features. 
Results of all the unsupervised methods are obtained using their official code with careful hyper-parameter tuning, with the same network architecture as \name.
Note that all the methods in this table are built on ImageNet-pretrained networks.
}
\label{tab:sota}
\vspace{-4mm}
\end{table*}

\vspace{-2mm}
\paragraph{Mining accuracy at training.}
We analyze the precision of positive mining along the training, which is defined as the ratio between the true positives mined and the total number of mined pseudo positives. 
We find that the precision of mining in both mini-batches and the memory bank increases gradually along the training.
The plots of the precision are displayed in the supplementary material.

\vspace{-2mm}
\subsection{Comparison with other methods}

We compare \name with the state-of-the-art methods in Table~\ref{tab:sota}, including the large-scale retrieval results by adding \dis.  
Following the convention in image retrieval and for a fair comparison, all the self-supervised methods (including SimCLR, MoCov2, SwAV \etc) start with ImageNet-pretrained networks and are used to fine-tune on GLDv2.

\vspace{-1mm}
\paragraph{Supervised methods.}
Table~\ref{tab:sota} shows that \name achieves outstanding mAP on \ox and \pa for both medium and hard setup. 
In particular, the mAP of \name on \ox medium is on par with the state-of-the-art supervised methods. Namely, only \emph{R101-GeM (GLDv2-clean)} performs better than \name on both \ox and \pa, when spatial verification (SP) is not considered.
In particular, with the same architecture, \name enhances the mAP on \ox medium/hard from 45.0/17.7 (ImageNet pretrained) to 73.1/48.3, comparing to 76.2/55.1 attained with full supervision.
Even with SP, \emph{DELF-R-ASMK+SP (GLD)} only performs better than \name when \dis is added. However, SP is much slower at run-time and memory-consuming.

\begin{table}[t]
\centering
{\small
\begin{tabular}{l c c c c }
Method & Oxf5k & Oxf105k & Paris6k & Paris106k  \\ \hline

\cite{iscen2018mining} &  78.2 & 72.6  &  85.1  & 78.0 \\ 
\name  &  92.0  & 91.4 &  94.2  & 90.5  \\

\end{tabular}
}
\vspace{-3mm}
\caption{\textbf{Evaluation on original Oxford5k and Paris6k.} 
}
\label{tab:mining}
\vspace{-1mm}
\end{table}
\begin{table}[t]
\centering
{\small
\begin{tabular}{l c c c}
Method & Labels & Validation set & Test set \\ \hline
\cite{weyand2020google} & Yes & 23.30 & 25.57 \\
ImageNet pretrained & No & 0.89 & 0.52 \\
InsCLR & No &13.39 & 13.71 \\

\end{tabular}
}
\vspace{-2mm}
\caption{\textbf{Retrieval task on GLDv2 (\% mAP@100).}}
\label{tab:gldv2}
\vspace{-5mm}
\end{table}

\begin{table}[t]
\centering
{\small
\begin{tabular}{l c c c}
Method & Fine-tuning & Labels & INSTRE  \\ \hline

ImageNet (w/o PCA) & - & - & 32.7 \\
\cite{iscen2017efficient}$^{\dagger}$   & Landmarks & Yes &  62.6 \\
\cite{iscen2018mining} &  INSTRE & No & 57.7 \\ 
\name w/o P.M. (\emph{nn}=1)  & INSTRE & No  & 55.6  \\
\name  & INSTRE   & No & 76.2 \\

\end{tabular}
}
\vspace{-3mm}
\caption{\textbf{Evaluation on INSTRE.}
$\dagger$ denotes the result of method \cite{gordo2017end} implemented by \cite{iscen2017efficient}. 
\emph{P.M.} denotes the positive mining in \name.
}
\label{tab:instre}
\vspace{-3mm}
\end{table}

\paragraph{Self-supervised methods.}
In the self-supervised regime, \name surpasses all the self-supervised methods.
We attempted to train SCAN on our task but it failed to converge in the clustering step (probably due to the large number of classes and significant intra-class variance). 
Lastly, although clustering-based methods like DeepCluster, PCL and SwAV implicitly take into account the intra-class variation into the representation learning, they hardly bring improvements.  It shows that learning intra-class invariance explicitly from positive and negative pairs is superior for the task at hand.
Moreover, as Table~\ref{tab:mining} shows, \name achieves significantly higher mAP than~\cite{iscen2018mining} on the original Oxford and Paris dataset.

\subsection{Evaluation on More Benchmarks}

\paragraph{GLDv2 retrieval task.} 
We directly evaluate the trained \name model on a recently released large-scale benchmark, namely, the GLDv2 retrieval task. As shown in Table~\ref{tab:gldv2}, InsCLR can achieve performance 13.39\% and 13.71\% on validation and test set, respectively. This is a surprisingly large improvement comparing to the ImageNet-pretrained baseline, given that no labels are used.

\paragraph{INSTRE benchmark.}
To showcase the generalization of \name, we fine-tune an ImageNet-pretrained ResNet-50 with GeM ($p=3$) on another instance retrieval benchmark: INSTRE~\cite{wang2015instre}.
We adopt the same train-test splitting as \cite{iscen2017efficient,iscen2018mining} for a fair comparison.
As shown by Table~\ref{tab:instre}, \name significantly outperforms \cite{iscen2018mining} and even \cite{iscen2017efficient} which uses labels and ResNet-101.
Moreover, the proposed positive mining within mini-batches and the memory again boosts the performance by a large margin (\ie 55.6 to 76.2).

\begin{figure}[t]
\centering
\includegraphics[width=0.99\columnwidth]{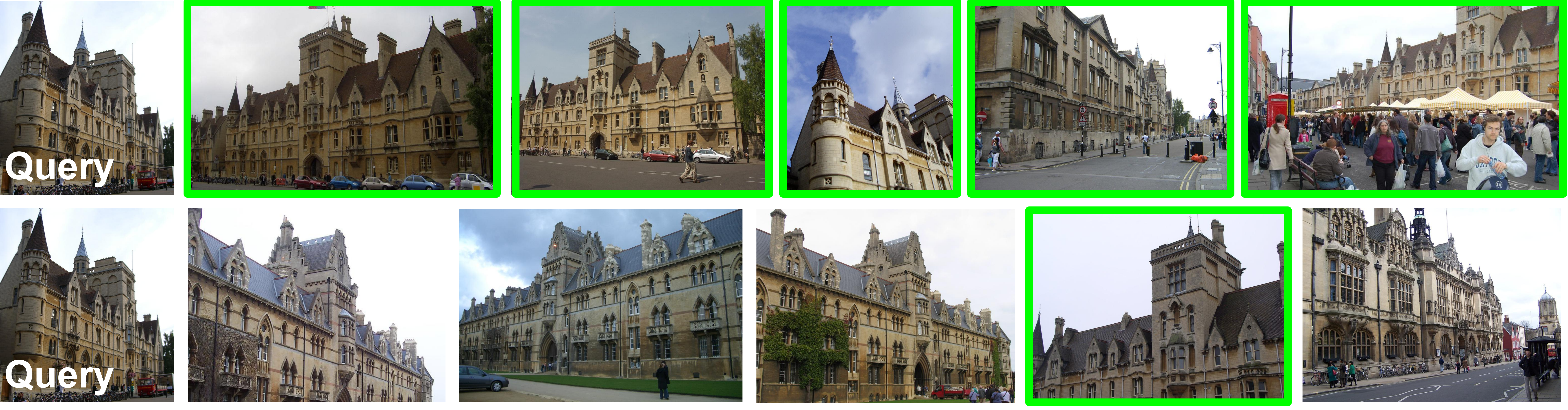}
\vspace{-2mm}
\caption{\textbf{\name} (top) \vs \textbf{MoCov2} (bottom).  
The top 5 retrieved images by \name are all correct (green). Some of them have very different viewpoint to the query image.
} 
\vspace{-6mm}
\label{fig:example}
\end{figure}

\vspace{-2mm}
\subsection{Qualitative Analysis}
\vspace{-1mm}
We further conduct qualitative analysis to demonstrate that \name can learn better representations for instance retrieval in terms of viewpoint-invariance than previous SSL methods. 
As we can see from Figure~\ref{fig:example}, \name can retrieve more correct images, and some of them have significantly different viewpoints with the query image.

\vspace{-2mm}

\section{Conclustion}
\vspace{-0mm}
We present  a new SSL method built on the instance-level constrastive learning for instance retrieval. This sets it apart from existing SSL methods that commonly learn from image-level contrast. \name can learn intra-class invariance by dynamically mining informative positives from both mini-batches and memory bank during training. Extensive experiments demonstrate that \name can achieve comparable performance to supervised methods on instance retrieval.

\bibliography{main.bbl}

\clearpage
\section{Supplementary Material}

\subsection{Relation to Set-based Retrieval Methods}
Other solution to tackle query-set-based image retrieval includes using the query expansion~\cite{chum2007total}. However, only global features (\ie no local features) are considered in our case, and thus the query expansion boils down to ranking the candidate pool using the averaged image features of the whole query set.
Another line of work that can improve a retrieval recall is to construct a similarity graph, and then explore the underlying image relation on manifolds using diffusion~\cite{iscen2017efficient,iscen2018mining} or graph traversal~\cite{chang2019explore}. Unfortunately, this line of approaches are usually computationally expensive in both offline (\eg the similarity graph construction) and online situations (\eg the graph traversal).

\subsection{Details of the Loss}
As mentioned in the main paper, we adopt a simple contrastive loss~\cite{hadsell2006dimensionality}. 
Concretely, the loss function $L$ for a training tuple can be computed as:

\begin{flalign}
\begin{split}
    L = \frac{1}{N_{p}}\sum_{i=1}^{N_{p}}\left [  \sum_{y_i \neq y_j}^{N_{iter}}S_{ij} - \sum_{y_i = y_j}^{N_{iter}}S_{ij}\right ],
\end{split}
\vspace{-2mm}
\end{flalign}

\noindent where $N_p$ is the number of selected pseudo positives plus an anchor image in a training tuple, defined as a `query set' for memory mining. Notice that a training mini-batch includes multiple training tuples, \eg, a mini-batch of 64 images contains 8 training tuples of size 8.
For a training iteration, $N_{iter}$ is a collection of all images in the mini-batch and the candidate pool for the anchor image. 
For the loss computation, the negatives include (1) all the remained images not selected as pseudo positives in the training tuple, (2) all the images in other training tuples in the same mini-batch, and (3) all the images in the candidate pool (excluding the mined positives).
$y_i \neq y_j$ indicates a negative pair while $y_i = y_j$ denotes a positive pair. 
$S_{ij}$ is the cosine similarity between the features of an image pair.
As described in the main paper, only negatives with a similarity higher than 0.4 (between its paired image) can contribute to the loss.

\subsection{Network Architecture in Detail}
Here we explain the network architecture adopted in this work in detail.
The network architecture  consists of three components: a backbone network, a spatial pooling layer and an embedding module. 
Given an input image, the backbone network is used to encode it to a feature map of size $C_b \times H \times W$, where $C_b$ is the number of channels and $H, W$ are the height and width of the feature map, respectively. The feature map is then fed into a spatial pooling layer to give a $C_b\times1$ dimensional feature vector. 
This feature vector is L2-normalized and passed to the embedding module which outputs a $C_e\times1$ dimensional feature. The embedding module acts as a similar role to PCA or whitening, but can be trained with the network in an end-to-end fashion.
Finally, the image representation is obtained by L2-normalizing the embedded feature.

Following~\cite{weyand2020google}, ResNet-101~\cite{he2016deep} is used as the backbone network, and only the first four convolutional blocks are kept. In this case, $C_b$ = 2048. 
The backbone is initialized with pretrained weights on ImageNet.
For the spatial pooling layer, we adopt the Generalized Mean pooling~\cite{radenovic2018fine} (GeM) with the parameter $p$ fixed to be 3. 
Similar to~\cite{gordo2017end,cao2020unifying}, we use a fully-connected layer as the embedding module with $C_e$ being 2048, without a careful tuning.

\begin{figure*}[ht!]
\centering
\includegraphics[width=0.85\linewidth]{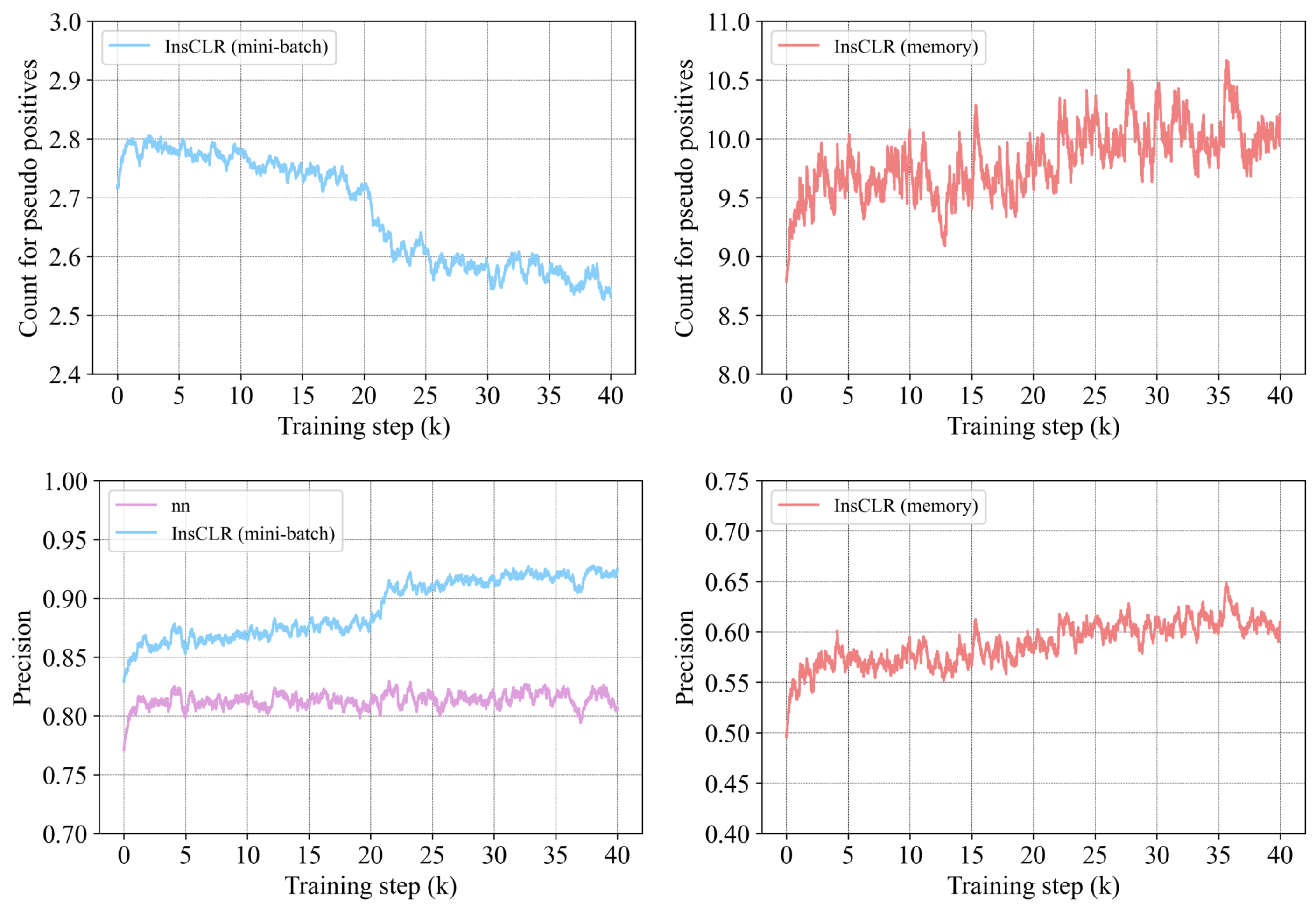}
\caption{ {Analysis of pseudo positives mined from the mini-batch (\textbf{left}) or memory bank (\textbf{right}) during training (of the second round).}
\textbf{Top}: the number of pseudo positives (for each anchor image) collected from mini-batches or memory bank.
\textbf{Bottom}: the precision of pseudo positives mined from the mini-batches and the memory bank.
Note that the precision is averaged over all the training tuples in each training iteration.
}
\vspace{-2mm}
\label{fig:analysis}
\end{figure*}

\subsection{Training Details}

\paragraph{Training \name.}
Data augmentations for training images include random cropping with a range of scale being 0.4 to 1.0 and aspect ratio being 0.75 to 1.33, followed by a resizing. The horizontal flipping with a chance of 50\% is also applied. 
The network is optimized using Adam~\cite{kingma2014adam}. 
The training of \name consists of two rounds of training, between which the candidate pool is updated. Namely, the candidate pool for the first round is obtained using the ImageNet-pretrained network, and that of the second round is obtained using the network at the end of the first round training.
For the first round, the augmented input images are resized to $224 \times 224$ with a batch size of 200.
For the second round, the augmented input images are resized to $448 \times 448$ with a batch size of 64. 
In this case, each mini-batch contains 16 training tuples with $N_b = 3$.
The size of the unaugmented images during training is fixed to be 512 $\times$ 512.
The training iteration of the first round is 60k with an initial learning rate of $10^{-4}$ and divided by 10 after 30k and 45k iterations. 
The training iteration of the second round is 40k, with an initial learning rate of $3 \times 10^{-5}$ and divided by 10 at 20k. 
The weight decay is $10^{-4}$.

\paragraph{Implementation details for other methods.}
All the existing methods are implemented using the official code or widely-used PyTorch implementations with careful hyper-parameter tuning.
Namely, MoCov2 is implemented using the official codebase~\footnote{https://github.com/facebookresearch/moco}.
PCL is implemented using the official codebase~\footnote{https://github.com/salesforce/PCL}.
SwAV is implemented using the official codebase~\footnote{https://github.com/facebookresearch/swav}.
DeepCluster, SimCLR and BYOL are implemented using the well-maintained repository~\footnote{https://github.com/open-mmlab/OpenSelfSup}.
For SimCLR, MoCov2, BYOL, PCL and SwAV, we follow the official training scheme, including the learning rate scheme and data augmentations \etc, and the network is trained for 20 epochs. We also try to train with these methods using different learning rates, and report with the best model.
The optimal initial learning rate of these five methods are 0.3, 0.03, 0.2, 0.03, 0.03, respectively.
For DeepClusters, we set the number of classes to be 30k, and train the network for 20 epochs with a learning of 0.15. 
The image pseudo labels and classifier are updated after each epoch.
All the methods are trained first trained with 224 image size and then 448, similar to that of \name.
The network of all the above methods is initialized using the ImageNet pretrained weights. This setup naturally results in a much faster convergence than that of the original setting (\ie starting from random initialization).
During training, the loss decreases gradually and the training is terminated when the loss flattens.

\subsection{Evaluation Details}
We compare different learning methods on three public benchmarks: Oxford and Paris with revisited annotations~\cite{radenovic2018revisiting} (denoted by \ox and \pa, respectively), and INSTRE~\cite{wang2015instre}. \ox has 4993 database images and \pa has 6322 database images, with 70 query images each. A more difficult evaluation is to add 1M distractor images \dis~\cite{radenovic2018revisiting} to the database, which can be view as a large-scale retrieval. 
INSTRE contains various objects including buildings, logos, toys \etc. Objects may have large variations in scale, view-points, and occlusions. INSTRE consists of 28,543 images, covering 250 different classes. We follow the evaluation protocol proposed in~\cite{iscen2017efficient}, which randomly splits the dataset into 1250 queries (5 per class) and 27293 database images. 

Ablation studies are conducted on the datasets without \dis, similar to~\cite{iscen2018mining,cao2020unifying}. The evaluation protocol is the mAP. For feature extraction at testing, we follow the convention~\cite{gordo2016deep,radenovic2018fine,cao2020unifying} to extract multi-scale representations by scaling the images with longest side being 1024 for \ox and \pa (and 256 for INSTRE) to three scales: \{$\sqrt{2}$, 1, $1/\sqrt{2}$\}, and average the features. The features of three scales are L2-normalized before and after taking the average.

\subsection{Quantitative Analysis on Training}

To better understand how the proposed mining of pseudo positives benefit the representation learning, we plot the number of the mined pseudo positives in mini-batches and the memory bank, as well as the precision of positive mining along the training.
The precision is defined as the ratio between the true positives mined and the total number of mined pseudo positives. 
\emph{Note that the ground truth image labels are used for computing the precision (only for analysis purpose), and are not involved in the training. }

As Figure~\ref{fig:analysis} shows, the number of selected pseudo positives in each mini-batch is initially around 2.8 and decreases to 2.5 towards the end of the training. Whereas in the memory bank, the number of pseudo positives is increased from 9 to 10.5, demonstrating that the model is able to mine more positives when  the model becomes more powerful during training.
We also compare our mining approach with \emph{nn} which takes all the images within a training tuple as positives. 
As shown in the bottom-left plot, the precision of pseudo positives mined by our method is consistently and significantly higher than  \emph{nn}, and increases along the training, reaching 92\% at the end.
On the memory side (the bottom-right plot), the precision is relatively low, since the most similar images are included in the training tuples within the mini-batch, and further mining a few true positives (might be hard positives) from the candidate pool of size 500 is very challenging. 
Nonetheless, the \name can still achieve a reasonable precision around 60\% (and attain 65\% at the highest), which benefits the training eventually.

\end{document}